\documentclass[letterpaper, 10 pt, conference]{ieeeconf}  % Comment this line out if you need a4paper

\IEEEoverridecommandlockouts                              % This command is only needed if 
\usepackage{graphics} % for pdf, bitmapped graphics files
\usepackage{graphicx}
\usepackage{epsfig} % for postscript graphics files
\usepackage{times} % assumes new font selection scheme installed
\usepackage{amsmath} % assumes amsmath package installed
\usepackage{amssymb}  % assumes amsmath package installed
\usepackage[OT1, T1]{fontenc}
\usepackage[hyphens]{url}
\usepackage{booktabs}
\usepackage[nospace]{cite}
\usepackage{multirow}
\usepackage{amsmath,amsfonts,bm}
\usepackage{svg}
\usepackage{tcolorbox}
\usepackage[hidelinks]{hyperref}
\usepackage{xurl}

\title{\LARGE \bf
Decoupled Early Exits for Task-Dependent Compute Allocation in Flow-Matching VLAs
}

\author{Riccardo Andrea Izzo$^{1}$, Rimvydas Rubavicius$^{2}$, Gianluca Bardaro$^{1}$,\\
Subramanian Ramamoorthy$^{2}$, Matteo Matteucci$^{1}$ and Alessandro Suglia$^{2}$% <-this % stops a space
\thanks{$^{1}$Riccardo Andrea Izzo, Gianluca Bardaro and Matteo Matteucci are with the
        Department of Electronics, Informatics, and Bioengineering, Politecnico di Milano, Milan, Italy
        {\tt\small \{riccardo.izzo, gianluca.bardaro, matteo.matteucci\}@polimi.it}. The work of R.A. Izzo was supported by the Italian Ministry of University and Research (MUR), funded by the European Union – NextGenerationEU (PNRR – M4C2, Inv. 3.3 – D.M. 630/2024), and co-funded by Oversonic Robotics s.r.l. Benefit Company (CUP D43C24001670008; Scholarship n. 40-033-16-DOT1316508-11141).}%
\thanks{$^{2}$Rimvydas Rubavicius, Subramanian Ramamoorthy and Alessandro Suglia are with the
        School of Informatics, University of Edinburgh, Edinburgh, UK
        {\tt\small \{rimvydas.rubavicius, s.ramamoorthy, asuglia\}@ed.ac.uk}. This project was supported by the Edinburgh International Data Facility (EIDF) and the Data-Driven Innovation Programme at the University of Edinburgh. The work of R. Rubavicius and S. Ramamoorthy was supported by UKRI Turing AI World Leading Researcher Fellowship on AI for Person-Centred and Teachable Autonomy under Grant (EP/Z534833/1).}%
}

\begin{document}

\bstctlcite{BSTcontrol}

\maketitle
\thispagestyle{empty}
\pagestyle{empty}

%%%%%%%%%%%%%%%%%%%%%%%%%%%%%%%%%%%%%%%%%%%%%%%%%%%%%%%%%%%%%%%%%%%%%%%%%%%%%%%%
\begin{abstract}
Flow-matching Vision-Language-Action (VLA) models have emerged as a potential solution for generalist robot control, designed by combining a pretrained Vision-Language Model (VLM) backbone with an action expert that generates continuous robot actions.
While these models exhibit impressive capabilities, due to their very high number of parameters, their computational requirements are often prohibitive for robotics control.
To mitigate these inefficiencies, existing methods predominantly skip VLM backbone layers with early exits or reduce denoising steps, while leaving action expert depth untouched.
We propose a framework that exposes backbone depth $V$, action expert depth $A$, and denoising steps $D$ as three jointly configurable compute axes in a VLA.
Starting from a pretrained VLA, we attach lightweight Exit Transformers (ET) at intermediate depths in both the backbone and the action expert, trained to distil the last layer of the policy into each exit.
Furthermore, we introduce a KV Cache synthesis mechanism that manages the missing keys and values of the skipped backbone layers, allowing the action expert to exit deeper than the backbone.
Finally, we show that the optimal compute budget is task-dependent, with different tasks benefiting from different axes and depths.
Notably, our method does not require training the original policy from scratch, and for each exit, it increases the number of parameters by only $2.1\%$ for SmolVLA and $4.1\%$ for $\pi_{0.5}$.
We validate our approach across two flow-matching VLAs (SmolVLA, $\pi_{0.5}$) and two benchmarks (LIBERO, Meta-World), revealing complementary effects: $V$ and $A$ respectively reduce FLOPs and latency, while $D$ improves both.
Our joint configurations $(V,A,D)$ reduce latency by $79.2\%$ and computation (FLOPs) by $31.8\%$, while improving mean success rate by $5.6\%$. 

\end{abstract}

\section{INTRODUCTION}
Vision-Language-Action (VLA) models map visual observations, language instructions, and current robot state directly to continuous control actions~\cite{kawaharazuka2025vla-survey,doi:10.1177/02783649261468360}.
VLAs are composed of a Vision-Language Model (VLM) that encodes the observations and instructions into a latent representation, and an action expert that generates a set of continuous actions through the denoising process conditioned on this latent representation. 
Modern state-of-the-art VLAs use a flow-matching objective for action generation to generate smooth and precise trajectories, iteratively refining the action predictions from noise~\cite{lipman2023flow}.

\begin{figure}[t]
    \centering
    \includegraphics[width=\linewidth]{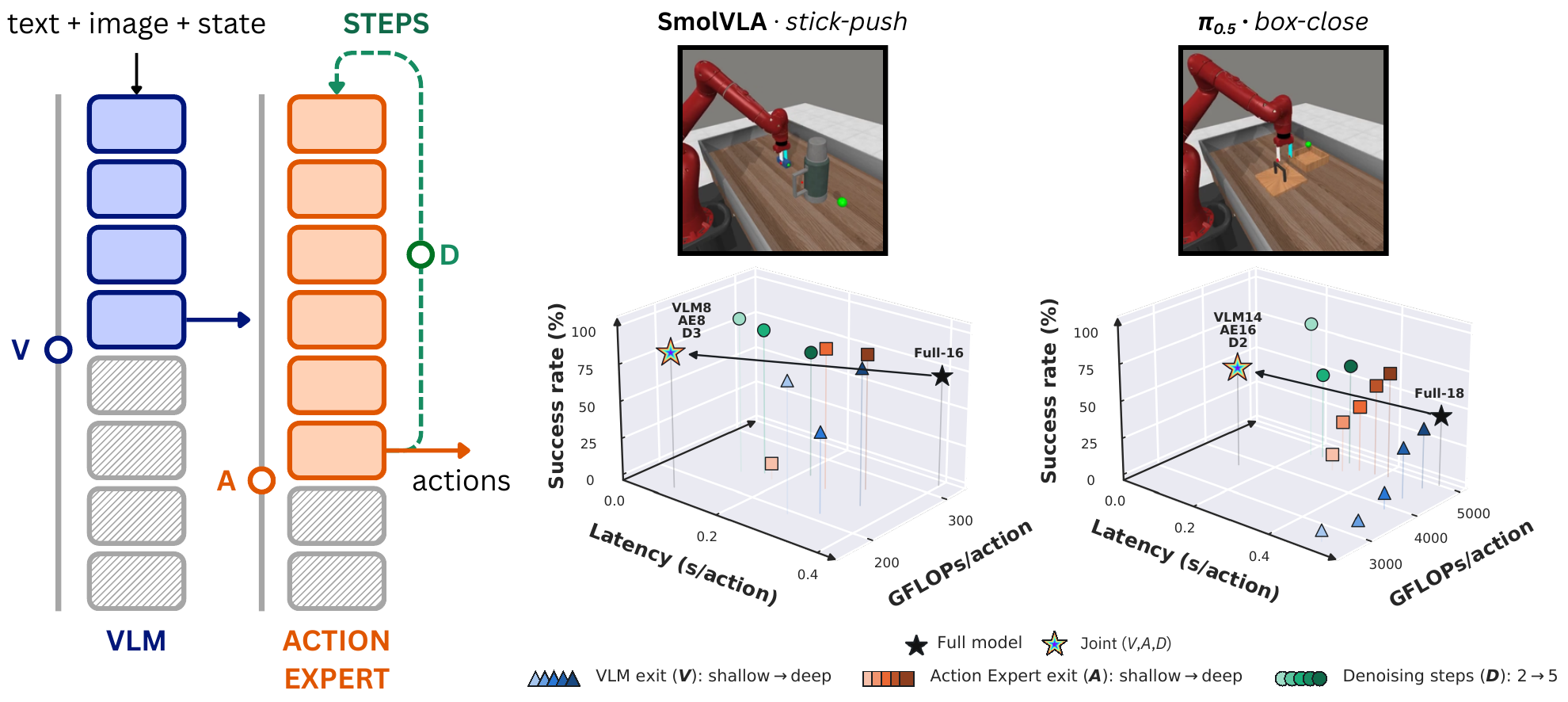}
    \caption{\textbf{$(V,A,D)$ compute axes on a flow-matching VLA.}
    Left: VLM depth $V$, the action expert depth $A$ and the denoising
    steps $D$. Right: multi-objective Pareto curves for two exemplary tasks on Meta-World MT50~\cite{yu2020meta}, \emph{stick-push} and \emph{box-close}, across SmolVLA and $\pi_{0.5}$.
    Our joint configuration outperforms the base policy in latency, FLOPs and success rate.}
    \label{fig:teaser}
\end{figure}

While VLAs exhibit increasingly strong generalisation capabilities, they remain computationally expensive, with inference times that are often impractical for real-time robot control frequency. 
This is particularly relevant for flow-matching VLAs, with the action head that requires multiple denoising steps to generate a single action chunk. 

\begin{figure}[t]
    \centering
    \includegraphics[width=1\linewidth]{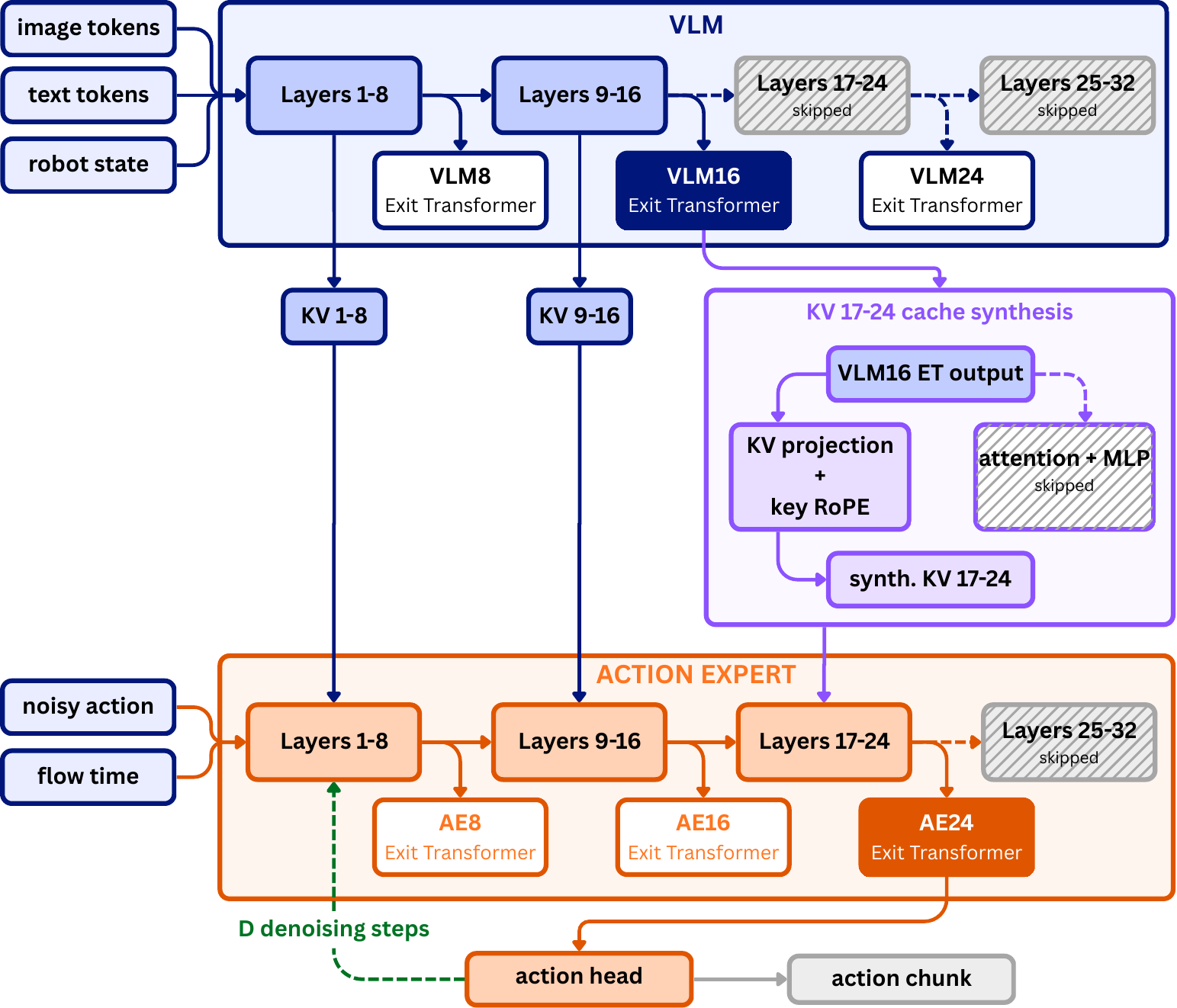}
    \caption{\textbf{Three-axis compute budget $(V,A,D)$.} Exit Transformers expose
    intermediate depths in the VLM ($V$) and the action expert ($A$), while $D$ sets the
    number of denoising steps. An early exit leaves the prefix KV cache of layers
    $V < i \le A$ empty, and those slots are synthesised from the ET output. Shown: SmolVLA with $(V,A,D)=(16,24,D)$.}
    \label{fig:architecture}
\end{figure}

To this end, previous methods have proposed different solutions to reduce the inference cost of VLAs, such as layer pruning~\cite{yang2026efficientvla,zhang2026mole}, early exits~\cite{yue2024deer,zhang2026a1}, or caching~\cite{xu2026vla}.
Almost all of them target the vision-language backbone, which holds most of the parameters and therefore dominates the FLOPs count, while others adaptively adjust the number of denoising steps~\cite{luan2026snapflow,zhang2026a1}.
Regarding this, the action expert has been largely overlooked, despite being the component that runs at every denoising step and therefore dominates latency.
To the best of our knowledge, many techniques have been proposed to reduce the \textit{depth} (i.e., the number of layers) of a vision-language backbone, but none has been applied to the action expert depth as an independent compute axis, and no study has measured the actual trade-off in how these axes interact.

We argue that early exit applies to the action expert as much as to the vision-language backbone, and that the two must be exposed jointly rather than used in isolation: early exiting the VLM backbone saves a cost paid once per action chunk, whereas exiting the action expert removes layers that would otherwise be executed at every denoising step.
We further show that the optimal budget is task-dependent, since different tasks have different semantics as well as different requirements in terms of precision and speed, and therefore the optimal compute budget may vary across tasks.
Fig.~\ref{fig:teaser} gives a glimpse of this on two Meta-World tasks with opposite requirements.
For \emph{stick-push} VLA requires more fine-grained dexterity, and is therefore limited by the action expert while tolerating a shallow backbone.
On the other hand, \emph{box-close} relies on perceiving the lid relative to the box, and is therefore limited by the VLM backbone while tolerating less depth in the action expert.

In this paper, we present a framework that exposes all three axes on a flow-matching VLA: backbone depth $V$, action expert depth $A$, and the number of denoising steps $D$.
Lightweight Exit Transformers (ET) are attached at intermediate depths of both the VLM backbone and the action expert.
Each ET is initialised as a copy of the transformer block at the corresponding depth, and is trained by distilling the last layer of the frozen policy into it, so that the pretrained action head is reused unchanged and every exit stays in the original action space.
The VLM backbone and the action expert are never updated, which preserves the exact behaviour of the full VLA and makes the optimal compute configuration a runtime choice rather than a separate model.
We also identified a bottleneck that arises when a backbone exit is used: the deeper expert layers still need the prefix keys and values that the skipped layers never computed, so we refill those cache slots by re-projecting the exited prefix with each skipped layer's own key and value matrices, at the cost of two projections instead of a full block. 
This further improves the savings of a backbone exit, without requiring additional training or inference cost.
Finally, the number of denoising steps ($D$) is the third axis and requires no training.
Fig.~\ref{fig:architecture} provides an overview of the framework.

Extensive evaluation shows that joint configurations reduce latency by $79.2\%$ and FLOPs by $31.8\%$, while improving mean success rate by $5.6\%$. On LIBERO and Meta-World benchmarks and across SmolVLA and $\pi_{0.5}$, our framework improves the Pareto frontier (success rate vs. latency/FLOPs) over full compute base models. %Further experiments on a SO-ARM 101 confirm the %these gains transfer on a physical platform.

Our contributions are:
\begin{itemize}
    \item A framework for flow-matching VLAs, with early exits on both the VLM backbone $V$ and the action expert $A$, paired with a KV cache synthesis mechanism and a variable number of denoising steps $D$, while adding a negligible number of parameters and avoiding training from scratch.
    \item An evaluation of the three axes on two flow-matching VLAs (i.e., SmolVLA and $\pi_{0.5}$) and two benchmarks (i.e., LIBERO and Meta-World), measuring trade-offs and the marginal effect of each against a full-compute model in terms of success rate, FLOPs, and latency.
    \item A task analysis, showcasing that different tasks benefit from different axes, and that the optimal compute budget is task-dependent.
\end{itemize}

For reproducibility, we release our code and models\footnote{\url{https://github.com/esgi-research-group/ee-vla}}.

\section{RELATED WORK}

\subsection{Vision-Language-Action Models}
Vision-Language-Action (VLA) models have emerged as a promising approach for generalist robotics policies by combining language, vision, and action prediction in an end-to-end framework. 
These policies are usually trained on massive multimodal datasets comprising images and language instructions paired with collected robot trajectories. 
Since the emergence of VLAs, early approaches such as RT-2~\cite{zitkovich2023rt} and OpenVLA~\cite{kim2024openvla} have built on top of pre-trained VLMs, processing image observations and text instructions to produce discretised actions. 
Due to the precision limitations of discretisation, subsequent models such as $\pi_0$~\cite{black2024pi0} transitioned to flow-matching networks to better capture the multimodal nature of robot trajectories. 
Furthermore, to address the computational requirements of previous VLAs, SmolVLA~\cite{shukor2025smolvla} and TinyVLA~\cite{wen2025tinyvla} proposed compact and efficient backbones without sacrificing performance.

\subsection{Efficient and Adaptive VLA Inference}
Among efficient inference approaches, one of the most effective and widely studied is early exiting~\cite{laskaridis2021adaptive}. This technique minimises computation by attaching intermediate predictors and terminating inference once a sufficiently reliable output is available. 
This paradigm was established for CNNs by BranchyNet~\cite{teerapittayanon2016branchynet}, and later extended for Transformers by DeeBERT~\cite{xin2020deebert}. 
Within the field of robotics, DeeR-VLA~\cite{yue2024deer} applied this concept to VLAs by adaptively terminating the VLM backbone based on the current situation or computational requirements. 
Concurrently, AC\textsuperscript{2}-VLA~\cite{yu2026ac} and MoLE-VLA~\cite{zhang2026mole} directly skip selected backbone layers. 
An orthogonal direction exploits temporal and structural redundancy: VLA-Cache~\cite{xu2026vla} reuses visual token features across control timesteps, while EfficientVLA~\cite{yang2026efficientvla} uses language layer pruning, visual token selection, and diffusion caching. 
Recent methods also target iterative action generation. 
For instance, CEED-VLA~\cite{song2025ceed} applies consistency distillation and early termination to the Jacobi decoding iterations of autoregressive VLAs. 
For flow-matching policies, SnapFlow~\cite{luan2026snapflow} distils generation to a single forward pass, while A1~\cite{zhang2026a1} exits when consecutive action predictions agree and uses inter-layer truncated flow matching to warm-start denoising at the next candidate depth. 
In A1, the backbone and the action expert are truncated together at the same layer, coupling their depth.
Our formulation instead exposes the backbone depth $V$ and action expert depth $A$ as independent axes through the use of decoupled early exits, alongside the number of denoising steps $D$. 
We argue that this decoupling matters because $V$ and $A$ have different computational roles, with the backbone being evaluated once per action chunk, while the action expert is evaluated at every denoising step. 
Our formulation takes advantage of a KV cache synthesis mechanism, allowing configurations with the action expert that exits deeper than the backbone ($A>V$). 
Moreover, while A1 is developed and trained around a specific Molmo-based architecture, our method is agnostic to flow-matching VLAs, and requires distilling early exits into frozen pretrained policies, thereby avoiding VLA pretraining.
We demonstrate this on two different VLAs (i.e., SmolVLA and $\pi_{0.5}$), and systematically characterise the individual and joint trade-offs of $V$, $A$, and $D$.

\section{METHOD}

\subsection{Preliminaries}
\label{sec:prelim}
Formally, the goal of a VLA is to learn a policy $\pi(\mathbf{A}_{t}\mid\mathbf{o}_t, l, \mathbf{q}_t)$ that, at each timestep $t$, maps a list of $k$ camera observations $\mathbf{o}_{t}=[o^{1}_{t}, \ldots, o^{k}_{t}]$, a language instruction $l$, and a proprioceptive state $\mathbf{q}_t \in \mathbb{R}^{d_q}$ (e.g., joint angles) to a chunk of $H$ future actions $\mathbf{A}_{t} = [a_{t}, a_{t+1}, \ldots, a_{t+H}]$, $a \in \mathbb{R}^{d_a}$.

\paragraph{Architecture}
Flow-matching VLAs realise $\pi$ with two components: a pretrained VLM backbone and an action expert, each implemented as a Transformer stack of $N$ layers indexed by $i \in \{1,\ldots,N\}$~\cite{shukor2025smolvla, intelligence2025pi05}.
At every layer, the token sequence splits into two contiguous blocks.
The \emph{prefix} carries the multimodal observation including the visual tokens extracted from $o_t$ by the VLM vision encoder, together with the tokenised instruction $l$, for a total of $M$ tokens.
The \emph{suffix} includes a linear projection of the proprioceptive state $\mathbf{q}_t$ and the $H$ noisy action tokens $\mathbf{A}^{\tau}_{t}$ at flow-matching time $\tau$.

We define $\mathbf{P}_{i} \in \mathbb{R}^{M \times d_{\mathrm{b}}}$ and $\mathbf{S}_{i} \in \mathbb{R}^{H \times d_{\mathrm{e}}}$ for the hidden states output by layer $i$ of the backbone and of the expert, respectively, where $d_{\mathrm{b}}, d_{\mathrm{e}}$ are the two model widths. When $i = 0$, $\mathbf{P}_{0}$ and $ \mathbf{S}_{0}$ correspond to the input embeddings.
The action expert is conditioned on the VLM backbone through the per-layer prefix keys and
values. Since the backbone and expert have the same depth, the suffix queries at
layer $i$ attend to the keys and values computed at that same layer from
$\mathbf{P}_{i-1}$.
The conditioning context is therefore the full set of per-layer prefix states, $\mathbf{c} \triangleq \{\mathbf{P}_{i}\}_{i=0}^{N-1}$, rather than a single pooled vector.
Following the default configurations of SmolVLA~\cite{shukor2025smolvla} and $\pi_{0.5}$~\cite{intelligence2025pi05}, we assume the backbone and expert have the same depth $N$, so that expert layer $i$ reads from backbone layer $i-1$.
At full depth, the action head $h$ maps the last $H$ suffix tokens of layer $N$ to the predicted velocity field, $v_{\theta}(\mathbf{A}^{\tau}_{t}, \tau, \mathbf{c}) \triangleq h(\mathbf{S}_{N})$.

\paragraph{Training and inference}
The policy is usually trained by imitation learning on a dataset of tuples $(\mathbf{o}_{t}, l, \mathbf{q}_{t}, \mathbf{A}_{t})$, minimising the discrepancy between predicted and ground-truth actions. Specifically, we use the conditional flow-matching loss~\cite{lipman2023flow}:
\begin{equation}
\mathcal{L}(\theta) = \mathbb{E}\big\lVert v_{\theta}(\mathbf{A}^{\tau}_{t}, \tau, \mathbf{c}) - u(\mathbf{A}^{\tau}_{t}\mid\mathbf{A}_{t})\big\rVert^{2},
\label{eq:fm}
\end{equation}
where, following the convention of $\pi_{0.5}$~\cite{intelligence2025pi05}, $\tau \in [0,1]$ is the flow-matching time, $\mathbf{A}^{\tau}_{t} = \tau\epsilon + (1-\tau)\mathbf{A}_{t}$ is the noisy chunk built from $\epsilon \sim \mathcal{N}(0,\mathbf{I})$, and $u(\mathbf{A}^{\tau}_{t}\mid\mathbf{A}_{t}) = \epsilon - \mathbf{A}_{t}$ is the target vector field.
At inference time, the chunk is generated by integrating the learned field from $\mathbf{A}^{1}_{t} \sim \mathcal{N}(0,\mathbf{I})$ down to $\tau = 0$ in $D$ Euler steps of size $\delta = 1/D$:
\begin{equation}
\mathbf{A}^{\tau-\delta}_{t} = \mathbf{A}^{\tau}_{t} - \delta\, v_{\theta}(\mathbf{A}^{\tau}_{t}, \tau, \mathbf{c}),
\label{eq:euler}
\end{equation}
after which $\mathbf{A}^{0}_{t}$ is executed on the robot.
Each of the $D$ steps re-runs the $N$ expert layers over the suffix, while $\mathbf{c}$ is computed once.
\subsection{Three-Axis Compute Budget}
\label{sec:ee}

We reduce inference cost along three axes and define the resulting compute budget as $(V,A,D)$.
$V$ and $A$ select the early exit depths of the VLM and action expert, respectively, while $D$ controls the number of denoising steps.
Accordingly, $\mathbf{P}_{V}$ is the prefix state at depth $V$ and $\mathbf{S}_{A}$ is the suffix state at depth $A$, with $V=A=N$ corresponding to full depth execution along both axes.

Our goal is to preserve success rate while reducing the number of layers executed, and therefore the FLOPs and latency of the policy.
However, since the original policy was trained to predict only from its final layer, intermediate states cannot be used directly, as the expert expects $\mathbf{P}_{N}$ and the action head $h$ expects $\mathbf{S}_{N}$.
Following FREE~\cite{bajpai2025free}, we implement $V$ and $A$ by attaching to each candidate depth a lightweight \emph{Exit Transformer} (ET), a single Transformer block initialised from the layer at that depth. 
Unlike FREE, we align the exits through direct supervision rather than adversarial training, as described in Section~\ref{sec:training}.

\paragraph*{\textbf{VLM ET ($V$)}} An early exit in the VLM consists of a Transformer layer whose architecture matches that of a layer in the VLM backbone.
At each candidate depth $V$, we attach an ET and apply it to the prefix with bidirectional self-attention.
The output of the ET, namely $\tilde{\mathbf{P}}_{V} = \mathrm{ET}_{V}(\mathbf{P}_{V})$, substitutes the layers that were not executed, so that the frozen action expert is conditioned on a representation approximating the prefix at full depth $\mathbf{P}_{N}$.

\paragraph*{\textbf{Action Expert ET ($A$)}} Similarly, an action expert ET consists of a single Transformer block that matches the architecture of the original action expert layer, using causal self-attention over the $H$ action tokens.
However, its objective is different from that of a VLM ET.
Instead of producing an intermediate representation that is consumed by another layer, it should directly yield the velocity from the suffix $\mathbf{S}_{A}$ produced at the selected depth $A$, $v_{\theta}^{(A)} = h(\mathrm{ET}_{A}(\mathbf{S}_{A}))$.
The remaining layers of the expert are never executed, saving a fraction $(N-A)/N$ of the suffix pass at each of the $D$ denoising steps.

\paragraph*{\textbf{Denoising steps ($D$)}} As defined in Eq.~\ref{eq:euler}, a denoising step is a single evaluation of the velocity field that advances the noisy chunk by $\delta$ towards the final trajectory. 
This is repeated $D$ times, defining the number of steps in the Euler integration.
Therefore, this third axis of our compute budget does not require any trainable parameters. 
The step count is a parameter that trades integration accuracy against compute and can be changed at inference time without retraining.

\subsection{KV Cache Synthesis}
\label{sec:kv}
At layer $i$, the action expert attends to the keys ($\mathbf{K}_i$) and values ($\mathbf{V}_i$) computed from the prefix at the corresponding layer $i$.
Both $\mathbf{K}$ and $\mathbf{V}$ are computed by the VLM backbone in a single prefix pass and stored in a cache that is later read at each denoising step.
A VLM early exit breaks this correspondence.
Building the prefix through depth $V$ only fills the KV cache slots up to that depth, so every action expert layer running deeper (i.e., $i > V$) finds its slot empty.
Prior works with LLMs either propagate the state of the exit layer to all deeper layers, projecting it with each layer's own key and value matrices~\cite{elbayad2020depth, schuster2022confident}, or recompute the missing entries when required~\cite{elhoushi2024layerskip}.
We follow state propagation~\cite{elbayad2020depth, schuster2022confident}, but propagate the ET output $\tilde{\mathbf{P}}_{V}$ instead of the prefix state $\mathbf{P}_{V}$ at the exit layer.
Every backbone layer $i$ with $V < i \le A$ projects it with its own key and value learnable matrices $\mathbf{W}_{k,i}$ and $\mathbf{W}_{v,i}$:
\begin{equation}
\resizebox{0.91\columnwidth}{!}{$\displaystyle
\mathbf{K}_{i} = \mathrm{RoPE}\big(\mathbf{W}_{k,i}\bar{\mathbf{P}}_{V}\big), \;
\mathbf{V}_{i} = \mathbf{W}_{v,i}\bar{\mathbf{P}}_{V}, \;
\bar{\mathbf{P}}_{V} = \mathrm{RMSNorm}_{i}(\tilde{\mathbf{P}}_{V})
$}
\label{eq:broadcast}
\end{equation}
so that the action expert finds the KV entries for all of its $A$ layers.
In practice, since the prefix is fixed for the whole action chunk and the outputs of the skipped layers are never read, we synthesise all the missing entries at once in the prefix pass, without running those layers.
As illustrated in Eq.~\ref{eq:broadcast}, each skipped layer only runs its RMSNorm, two projections, and key rotation, while its attention, output projection, and feed-forward are skipped, leading to significant speedups outlined in Table~\ref{tab:kvcache}.
This synthesis adds no approximation beyond state propagation itself, as the synthesised entries are identical to those obtained by running each skipped layer on $\tilde{\mathbf{P}}_{V}$.
At the end, the prefix pass costs $V$ layers, the ET and the projections of Eq.~\ref{eq:broadcast}, instead of $N$ full layers.

\begin{table}[t]
\centering
\caption{Training hyperparameters of the ET for each checkpoint.}
\label{tab:hyperparams}
\scriptsize
\resizebox{\columnwidth}{!}{%
\begin{tabular}{lcccccc}
\toprule
Checkpoint & $N$ & Exits & LR & Warm. & Batch & Steps \\
\midrule
\textit{HuggingFaceVLA/smolvla\_libero}        & $32$ & $\{8,16,24\}$ & $10^{-4}$ & $1000$ & $8$                       & $31{,}184$ \\
\textit{lerobot/smolvla\_metaworld}            & $16$ & $\{4,8,12\}$  & $10^{-4}$ & $1000$ & $8$                       & $25{,}600$ \\
\textit{lerobot/pi05\_libero\_finetuned\_v044} & $18$ & $\{6,9,12,14,16\}$ & $10^{-4}$ & $500$  & $8$                       & $15{,}592$ \\
\textit{tiantianx/pi05\_metaworld}             & $18$ & $\{6,9,12,14,16\}$ & $10^{-4}$ & $1000$ & $8$                       & $25{,}600$ \\
\bottomrule
\end{tabular}%
}
\end{table}

\subsection{Training}\label{sec:training}

We train the early exits by distillation, with the last layer of the corresponding module as teacher ($\mathbf{P}_{N}$ for the VLM backbone and $\mathbf{S}_{N}$ for the action expert) and each ET as student, thus aligning their representations.
The backbone and the action expert are never updated, so that the behaviour of the original policy at full compute is not affected.
We denote by $v_{\theta}$ the full-depth velocity, and by $v_{\theta}^{(V)}$ and $v_{\theta}^{(A)}$ the velocity when the backbone exits at depth $V$ or the expert at depth $A$, respectively.
Since the action expert and VLM ETs produce different outputs, the two are supervised by different training objectives.

\paragraph*{\textbf{VLM ET}} A VLM ET is trained to output an intermediate representation that every action expert layer deeper than $V$ will consume. 
The loss for each VLM exit at depth $V$ minimises:
\begin{equation}
\mathcal{L}_{V} = \lambda_{a}\,\mathrm{MSE}\big(v_{\theta}^{(V)}, v_{\theta}\big)
+ \lambda_{d}\, d\big(\tilde{\mathbf{P}}_{V}, \mathbf{P}_{N}\big).
\label{eq:loss_vlm}
\end{equation}
The first term is a velocity distillation that minimises the Mean Squared Error (MSE) between $v_{\theta}^{(V)}$ and $v_{\theta}$.
The second term is a feature distillation that instead minimises the cosine distance $d$ between the prefix $\tilde{\mathbf{P}}_{V}$ and the one at full depth $\mathbf{P}_{N}$.
The first term, therefore, measures the effect of the intermediate representation on the generated action, while the second supervises the alignment of the representations.

\paragraph*{\textbf{Action expert ET}} 
An action expert ET is instead trained to output the velocity directly.
Since no deeper layer consumes its representation, the feature distillation term is dropped in favor of a standard flow-matching loss term.
The loss for each action expert exit at depth $A$ minimises:
\begin{equation}
\mathcal{L}_{A} = \lambda_{f}\,\mathrm{MSE}\big(v_{\theta}^{(A)}, u\big)
+ \lambda_{a}\,\mathrm{MSE}\big(v_{\theta}^{(A)}, v_{\theta}\big).
\label{eq:loss_ae}
\end{equation}
The first term is the flow-matching objective of Eq.~\ref{eq:fm}, while the second term is the same distillation term as in Eq.~\ref{eq:loss_vlm}.

We set $\lambda_{f}$, $\lambda_{a}$ and $\lambda_{d}$ with a grid hyperparameter search, retaining the configuration with the highest success rate on LIBERO and Meta-World at the corresponding exit depths.
This yields $\lambda_{a} = 1$, $\lambda_{d} = 0.2$ for Eq.~\ref{eq:loss_vlm} and $\lambda_{f} = 1$, $\lambda_{a} = 0.5$ for Eq.~\ref{eq:loss_ae}.
Exits at different depths are trained jointly for each axis by averaging their losses, so that the objective does not depend on the number of enabled exits, and each ET receives gradients only from its own term.
Each ET is trained against a full depth counterpart on the other axis, and the two are composed only at inference time.
Finally, the addition of the ETs does not significantly increase the overall model size. 
For SmolVLA, an action expert ET adds $3.1$M parameters and a VLM ET adds $9.8$M, against the $605$M of the frozen model. 
For $\pi_{0.5}$, the same ETs add $23.6$M and $110.1$M parameters to a total of $3.3$B. 

\begin{figure*}[t]
    \centering
    \includegraphics[width=0.96\textwidth]{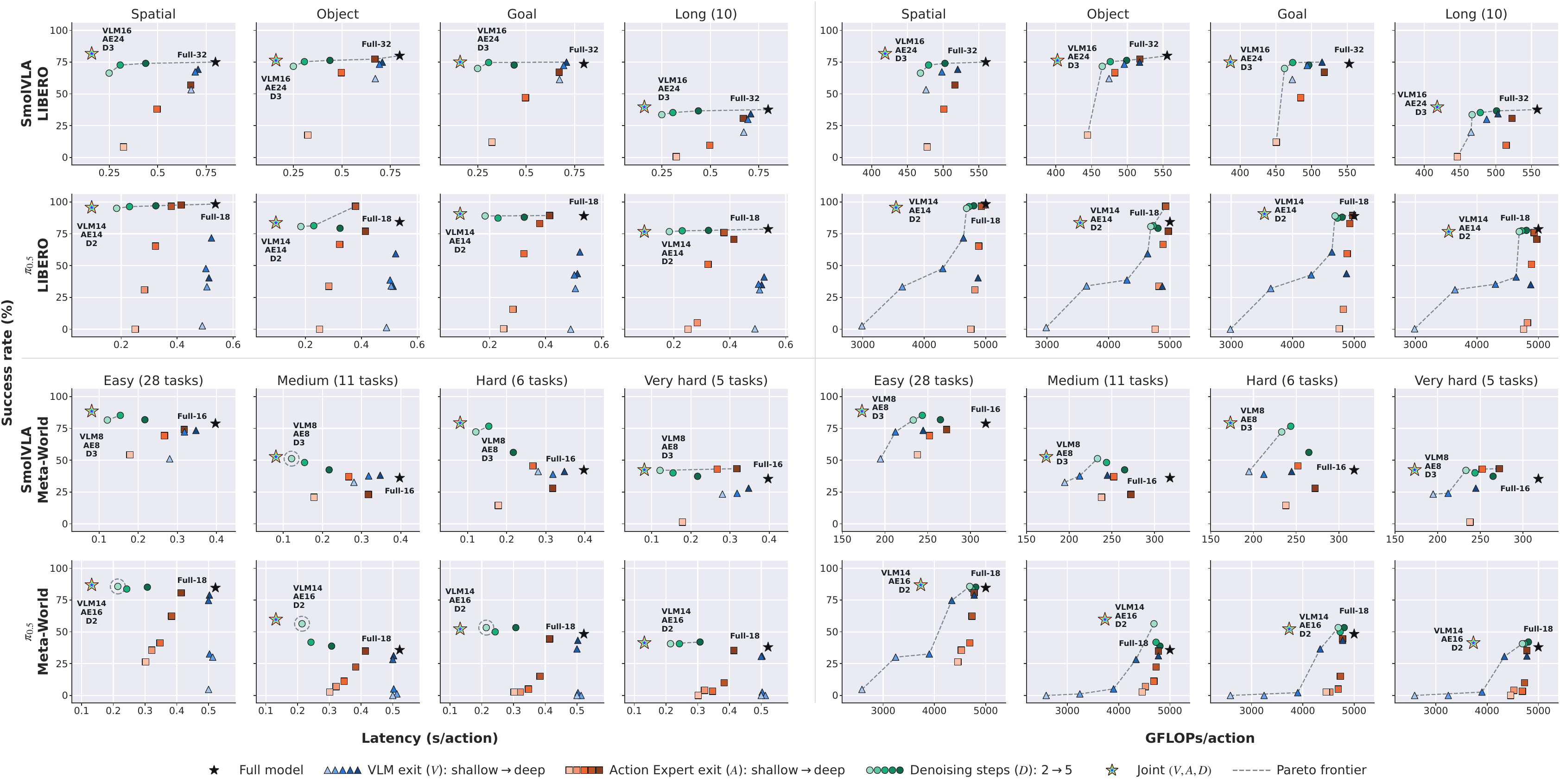}
    \caption{\textbf{Pareto curves on three axes.} Success rate (SR) against latency (left) and
    FLOPs (right) for SmolVLA and $\pi_{0.5}$, across LIBERO (top) and Meta-World (bottom). Each point changes one axis of the compute budget
    $(V, A, D)$ while the other two are fixed, with the full policy
    as reference. All VLM exits ($V$) points include the KV cache synthesis of
    Eq.~\ref{eq:broadcast}; rainbow stars denote the optimal joint configurations $(V, A, D)$ reported in
    Table~\ref{tab:joint_configs}. Every point is averaged over 30 episodes per task, so
    $N=300$ per LIBERO suite and $N=840/330/180/150$ for the four Meta-World difficulty
    groups, which corresponds to a mean 95\% confidence interval of $\pm2.67\%$ on
    LIBERO and $\pm3.33\%$ on Meta-World.}
    \label{fig:pareto}
\end{figure*}

\section{EXPERIMENTS}
\label{sec:results}

\subsection{Experimental Setup}
\label{sec:setup}

\paragraph*{\textbf{Benchmarks}} 
We evaluate our method on two benchmarks: LIBERO~\cite{liu2023libero}, a large-scale benchmark for long-horizon robot manipulation, and Meta-World \textit{MT-50}~\cite{yu2020meta}, a suite of 50 robotic manipulation tasks across four levels of difficulty (i.e., easy, medium, hard, and very hard).

\paragraph*{\textbf{Datasets and models}} 
We train the ETs within the LeRobot~\cite{cadene2026lerobot} framework on the datasets corresponding to the benchmarks, namely LIBERO (\textit{lerobot/libero}) and Meta-World (\textit{lerobot/metaworld\_mt50}).
For both SmolVLA and $\pi_{0.5}$, we start from a checkpoint already fine-tuned on the dataset, which is kept frozen to train only the ETs.
We report all the hyperparameters in Table~\ref{tab:hyperparams}.

\paragraph*{\textbf{Evaluation Metrics}}
We report success rate (SR), the fraction of successful episodes over all tasks of a suite; FLOPs, measured with the PyTorch profiler on one action-generation call (backbone plus the full denoising loop); and latency, the wall-clock time of that call, averaged over the evaluation.
Latency is measured at batch size one to avoid batched throughput.
All training and evaluation runs are performed on a single NVIDIA A100 with 40\,GB of VRAM.

Specifically, we aim to address the following research questions (RQ):
\begin{itemize}
    \item \textbf{RQ1 (Axis roles and composition):} How do $V$, $A$ and $D$ individually trade success rate against latency and FLOPs, and can their complementary effects be composed?
    \item \textbf{RQ2 (Task dependence):} Is the optimal compute budget task-dependent? Do different tasks and policies favour different axes and depths?
    \item \textbf{RQ3 (Early Exits):} Can early exits be applied to the action expert and are they beneficial over layer truncation? Is the KV cache synthesis effective in allowing the action expert to run deeper than the backbone?
\end{itemize}

\begin{table}[t]
\centering
\caption{\textbf{Base and joint configurations.} Joint configurations $(V, A, D)$ outperform the full compute baseline on average in success rate, latency, and FLOPs.}
\label{tab:joint_configs}
\scriptsize
\setlength{\tabcolsep}{3pt}
\resizebox{\columnwidth}{!}{%
\begin{tabular}{lccccc}
\toprule
Model / Benchmark & Version & $(V,A,D)$ & SR (\%) & Latency (s) & GFLOPs \\
\midrule
\multirow{2}{*}{SmolVLA / LIBERO}     & Base  & $(32,32,10)$ & $66.6$          & $0.798$          & $556.6$ \\
                                       & Joint & $(16,24,3)$ & $\mathbf{68.2}$ & $\mathbf{0.159}$ & $\mathbf{406.3}$ \\
\cmidrule(lr){1-6}
\multirow{2}{*}{SmolVLA / Meta-World} & Base  & $(16,16,10)$ & $60.7$          & $0.397$          & $318.0$ \\
                                       & Joint & $(8,8,3)$   & $\mathbf{74.9}$ & $\mathbf{0.081}$ & $\mathbf{173.0}$ \\
\midrule
\multirow{2}{*}{$\pi_{0.5}$ / LIBERO}     & Base  & $(18,18,10)$ & $\mathbf{87.6}$          & $0.537$          & $5000.2$ \\
                                           & Joint & $(14,14,2)$ & $86.7$          & $\mathbf{0.095}$ & $\mathbf{3541.0}$ \\
\cmidrule(lr){1-6}
\multirow{2}{*}{$\pi_{0.5}$ / Meta-World} & Base  & $(18,18,10)$ & $64.9$          & $0.522$          & $5000.2$ \\
                                           & Joint & $(14,16,2)$ & $\mathbf{72.2}$ & $\mathbf{0.132}$ & $\mathbf{3733.2}$ \\
\bottomrule
\end{tabular}
}
\end{table}

\begin{figure*}[t]
    \centering
    \includegraphics[width=\linewidth]{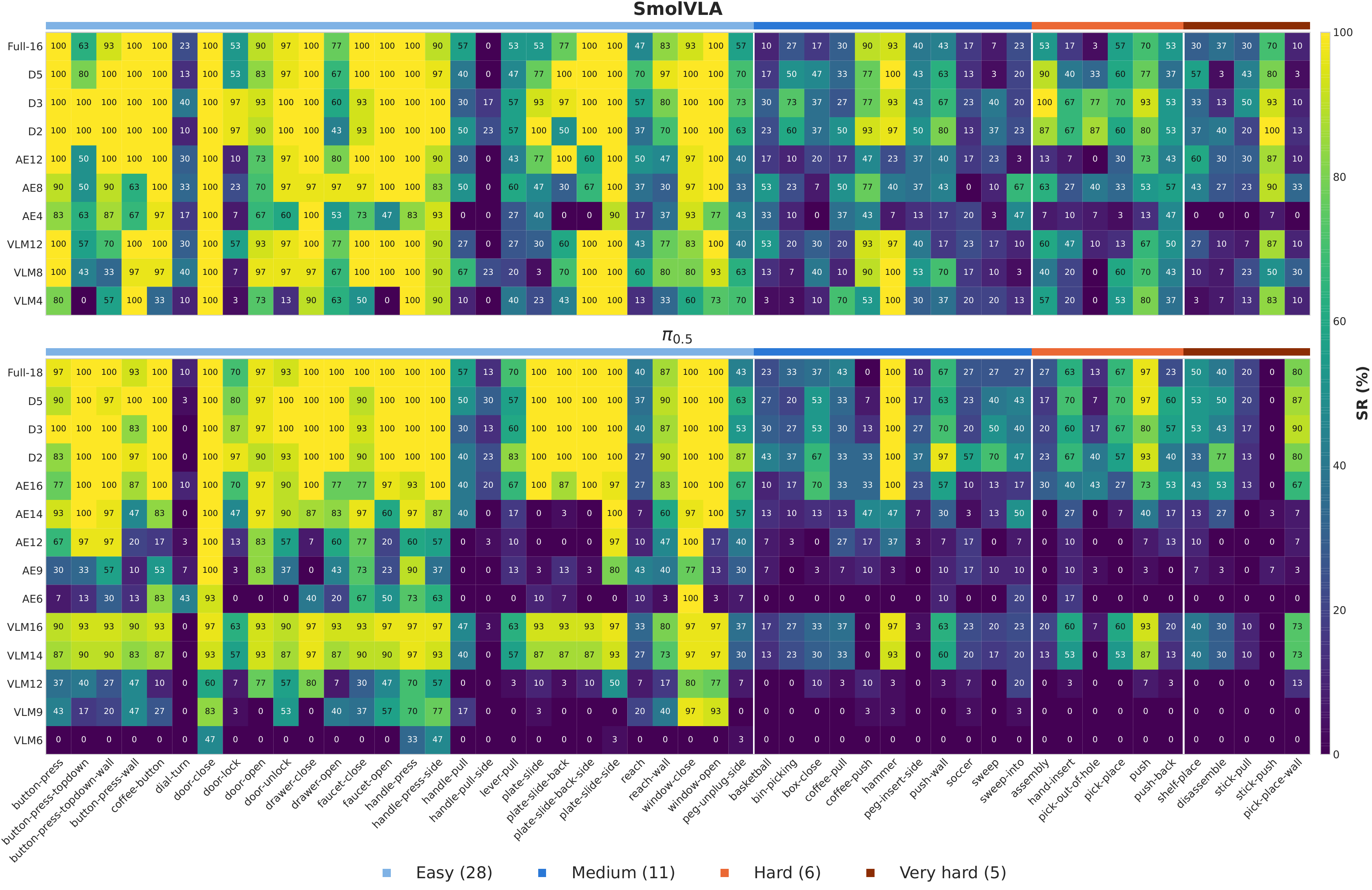}
    \caption{\textbf{Task analysis with the three axes on Meta-World.}
    Success rate (SR) for SmolVLA (top) and $\pi_{0.5}$ (bottom) across
    compute budget configurations (rows) and tasks (columns). Rows    
    compare the full baseline (Full) with different compute budgets in $(V, A, D)$, where only one axis is modified at a time.
    Each point is averaged over 30 episodes per task.
    The full policy is optimal on only 11/50 tasks for SmolVLA and 22/50 for $\pi_{0.5}$, with the best configurations distributed across all three axes. 
    Tasks have different requirements and benefit from different axes, motivating task-dependent $(V,A,D)$.}
    \label{fig:metaworld_heatmaps}
\end{figure*}

\subsection{Single-Axis Trade-offs and Joint Composition (RQ1)}
\label{sec:pareto}
% ANSWER: role of each axis
\paragraph*{\textbf{Individual axis roles}}
In Fig.~\ref{fig:pareto} we analyse the trade-offs between success rate and both latency and FLOPs, for each axis of the compute budget $(V, A, D)$.
We observe that the three axes have different effects on the two metrics, with $V$ being the most effective on FLOPs, $A$ on latency and $D$ on both.
The number of denoising steps $D$ appears to be a free lever in terms of both latency and FLOPs, usually with a negligible effect on success rate, while sometimes substantially improving performance on Meta-World with $\pi_{0.5}$.
This suggests that the optimal number of denoising steps is often not the default one, and that simply reducing it can improve both efficiency and performance, creating a single point on the Pareto curve that dominates the full policy.
The action expert early exit $A$ has a strong effect on latency, often halving it when using shallower exits while preserving success rate.
On the other hand, the effect on the FLOPs is more contained. 
Smaller policies such as SmolVLA also have a modest reduction in FLOPs up to approximately $20\%$, while larger ones such as $\pi_{0.5}$ simply retain the full performance.
Regarding the VLM early exit $V$, we observe substantial FLOPs reductions across both policies and benchmarks.
Contrarily to the other two axes, the effect of $V$ on latency is more contained.
% However, $V$ appears to be the most effective axis on FLOPs, with consistent gains across both policies and benchmarks.
Finally, this highlights the complementary nature of the three axes, with each one being more effective on a different metric.

\paragraph*{\textbf{Joint configurations}}
For each model and benchmark pair, we select the operating point with the best trade-off in terms of success and efficiency on each axis and compose the resulting choices of $V$, $A$, and $D$.
Table~\ref{tab:joint_configs} compares these optimal joint configurations with their corresponding full compute base policies.
Across the four settings, the joint configurations reduce latency by $79.2\%$ and FLOPs by $31.8\%$ on average, while improving mean success rate by $5.6\%$.
This confirms that the complementary gains of the three axes can be composed into a single joint configuration that improves the success/efficiency trade-off over the full policy, as well as the corresponding single axis choices.

\begin{figure}[t]
    \centering
    \includegraphics[width=1\linewidth]{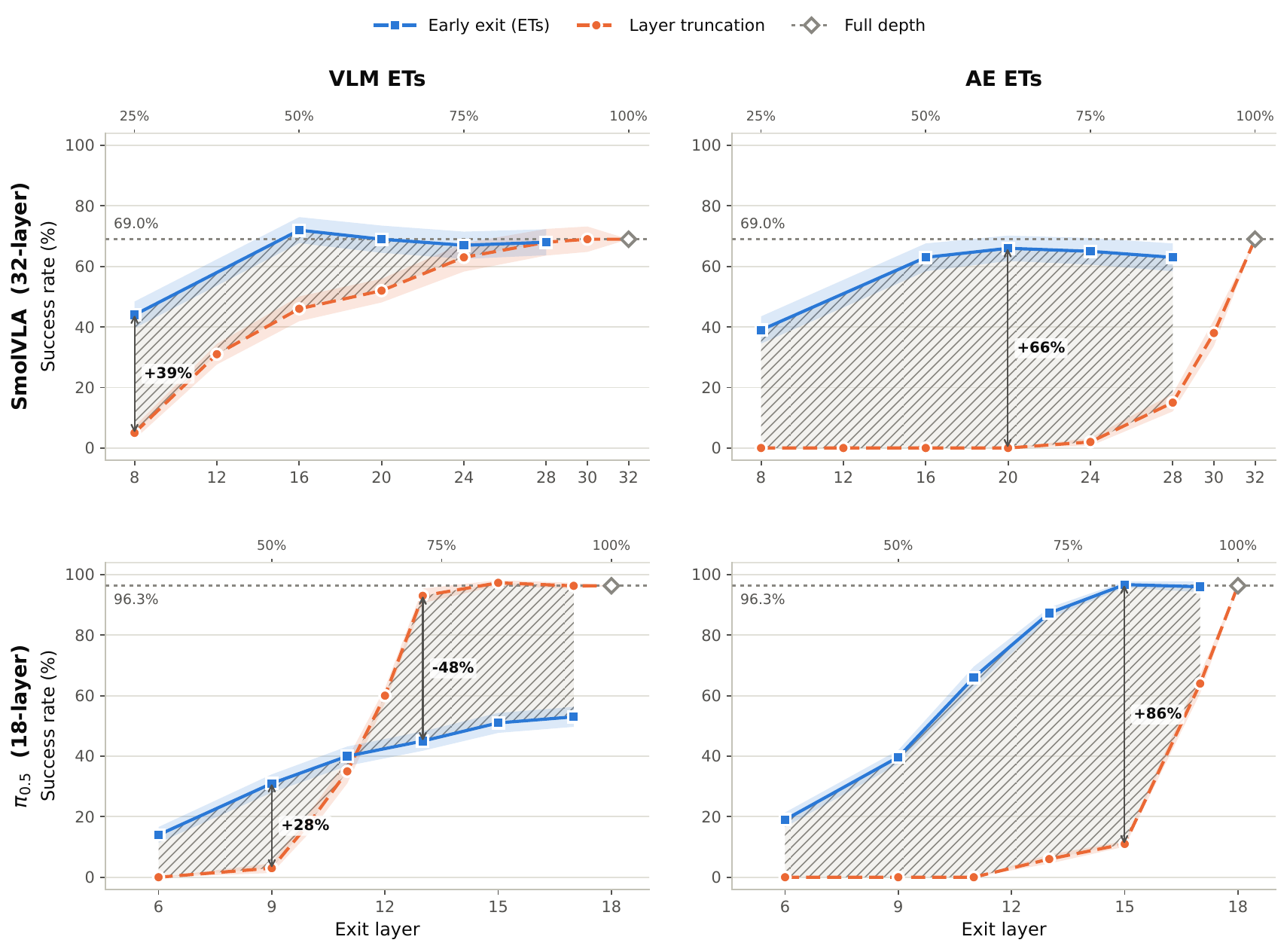}
    \caption{\textbf{Early exits vs. layer truncation.} Average success rate on LIBERO
    as a function of the exit layer, for the VLM axis $V$ (left) and the action expert
    axis $A$ (right), on SmolVLA (top) and $\pi_{0.5}$ (bottom).
    \emph{Layer truncation} reads the action head directly from the intermediate state,
    while \emph{early exit} routes it through an ET at that depth. The dashed line is the
    full depth policy; arrows report the gap at the corresponding depth.}
    \label{fig:exit_vs_truncation}
\end{figure}

\subsection{Task-Dependent Compute Requirements (RQ2)}
\label{sec:task_analysis}

In this section, we investigate the relation between different types of tasks and each axis in our compute budget $(V, A, D)$. The results in Section~\ref{sec:pareto} identify one useful operating point, but implicitly assume that every task can save computation from the same part of the policy.
We test this assumption with SmolVLA and $\pi_{0.5}$ on Meta-World MT50, whose 50 tasks range across four levels of difficulty. The question is therefore not only \emph{how much} compute a task needs, but \emph{where} to allocate compute in the policy.

\paragraph*{\textbf{A fixed compute budget is suboptimal across tasks}} The full policy is the best configuration on only $11/50$ tasks for SmolVLA and $22/50$ tasks for $\pi_{0.5}$, with the optimal configuration being spread across the three axes. 
The selection of the best axis for each task would raise mean success from $60.7\%$ to $79.5\%$ on SmolVLA and from $64.9\%$ to $75.9\%$ on $\pi_{0.5}$.
For instance, for task \textit{stick-push} with SmolVLA, the optimal compute configuration for this very hard suite would have been $(8, 8, 3)$, while $(4, 8, 2)$ could achieve even more significant speedups while preserving the success rate.
This task-dependent compute budget shows that the compute allocation becomes a choice of \emph{where} to allocate compute across perception $V$, action generation $A$, and denoising $D$, rather than only \emph{how much} compute to use.

\paragraph*{\textbf{Tasks are sensitive to different compute axes}} 
The heatmaps in Figure~\ref{fig:metaworld_heatmaps} do not degrade uniformly with depth, as a task that can tolerate a shallow exit on one axis can fail on a different one.
For example, on SmolVLA, \emph{hammer} retains $100\%$ SR at VLM4 but falls from $93\%$ to $7\%$ at AE4, whereas \emph{faucet-open} retains $100\%$ at AE8 but falls from $100\%$ to $0\%$ at VLM4.
In particular, tasks whose goal region tolerates a coarse trajectory (e.g., \emph{door-close}, \emph{drawer-close}, \emph{window-open/close}, \emph{handle-press}) retain a high success rate even with shallow exits in both policies, while tasks that require precise contact collapse (e.g., SmolVLA AE8 on \emph{hammer} $93\!\to\!40$,
\emph{reach-wall} $83\!\to\!30$; $\pi_{0.5}$ AE14 on \emph{pick-place} $67\!\to\!7$). This suggests that the action expert depth $A$ matters when that representation must be converted into a precise trajectory, hence for tasks requiring more dexterity. On the other hand, the VLM depth $V$ is more effective when the bottleneck is the visual complexity of the scene. For instance, on SmolVLA, shallow exits are harmless for large and unambiguous targets (e.g., \emph{door-close}, \emph{plate-slide-side} and \emph{handle-press}) that retain $100\%$ with only VLM4, but fatal for small affordances and obstacle variants (e.g., \emph{faucet-open} drops $100\!\to\!0$, \emph{coffee-button} $100\!\to\!33$).
Finally, among the three axes, the number of denoising steps $D$ is the most robust axis, as it improves the majority of tasks. Specifically, it improves tasks that require more dexterity and where the full policy fails, such as SmolVLA's Hard group from $42\%$ to $77\%$ (e.g., \emph{pick-out-of-hole} $3\!\to\!77$ and \emph{assembly} $53\!\to\!100$). 

For a robot executing different manipulation tasks in the real world, compute could potentially be treated as a control variable rather than a fixed property of the policy. The robot can allocate capacity to the component that most benefits the current task, reserving depth in the backbone for complex semantic domains and action expert depth when dexterity is required. Our results establish potential headroom for task-dependent compute allocation, with the online selection of $(V, A, D)$ based on the type of tasks that remains an important direction for future work.

\begin{table}[t]
\centering
\caption{\textbf{KV cache synthesis ablation.} Reported gains of the KV cache synthesis mechanism over the standard approach on
full LIBERO (300 episodes per suite for SmolVLA, 400 for $\pi_{0.5}$), for each VLM early exit $V$.}
\label{tab:kvcache}
\scriptsize
\setlength{\tabcolsep}{3pt}
\resizebox{\columnwidth}{!}{%
\begin{tabular}{lccc@{\hskip 8pt}ccccc}
\toprule
& \multicolumn{3}{c}{SmolVLA (32 layers)} & \multicolumn{5}{c}{$\pi_{0.5}$ (18 layers)} \\
\cmidrule(lr){2-4}\cmidrule(lr){5-9}
$V$ & 8 & 16 & 24 & 6 & 9 & 12 & 14 & 16 \\
\midrule
VLM GFLOPs w/o   & 86.3 & 86.3 & 86.3 & 3939 & 3939 & 3939 & 3939 & 3939 \\
VLM GFLOPs w/    & \textbf{21.6} & \textbf{43.1} & \textbf{64.7} & \textbf{1094} & \textbf{1970} & \textbf{2845} & \textbf{3283} & \textbf{3648} \\
Speedup ($\times$)            & 4.0 & 2.0 & 1.3 & 3.6 & 2.0 & 1.4 & 1.2 & 1.1 \\
\midrule
VLA GFLOPs w/o    & 559 & 559 & 559 & 5213 & 5213 & 5213 & 5213 & 5213 \\
VLA GFLOPs w/     & \textbf{495} & \textbf{516} & \textbf{538} & \textbf{2368} & \textbf{3244} & \textbf{4119} & \textbf{4557} & \textbf{4894} \\
Speedup (\%)       & $+11.6$ & $+7.7$ & $+3.9$ & $+54.6$ & $+37.8$ & $+21.0$ & $+12.6$ & $+6.1$ \\
\midrule
$\Delta$SR (\%)     & $+0.5$ & $-1.8$ & $+1.8$ & $0.0$ & $0.0$ & $+1.0$ & $-0.5$ & $0.0$ \\
\bottomrule
\end{tabular}%
}
\end{table}

\subsection{Ablation Studies (RQ3)}
\label{sec:ablations}

\subsubsection{Exit Transformers vs. Layer Truncation}
The simplest way to reduce depth is layer truncation, which stops the forward pass at an intermediate layer and uses the resulting representation without additional parameters or training. 
We compare our ETs against this baseline to assess whether trained exits improve success rate enough to justify their additional parameters and computation, particularly at shallow depths.
In our implementation of layer truncation, the action head reads directly from the intermediate layer representation without an ET.
Fig.~\ref{fig:exit_vs_truncation} shows the results of this comparison. In the action expert, the early exits consistently outperform layer truncation, especially at shallower depths, where the gap can be as large as 86\% in success rate.
Regarding the VLM, we notice the same results on SmolVLA, while for $\pi_{0.5}$ we observe a different trend: early exits perform better at shallow depths, but at deeper ones $\pi_{0.5}$ recovers and outperforms the early exits. 
We attribute this to the larger scale of $\pi_{0.5}$, whose intermediate representations are already informative and appear to be sufficient to condition the action head without the need for an ET. 
Consistently, in the action expert, where the size gap between the two models is smaller, the ETs are effective at all evaluated depths in both models.

\subsubsection{KV Cache Synthesis Ablation}
In this section, we aim to isolate the effect of our KV cache synthesis mechanism.
Table~\ref{tab:kvcache} illustrates the gains on LIBERO with SmolVLA and $\pi_{0.5}$, for each VLM early exit $V$ selected in Table \ref{tab:hyperparams}.
On both VLAs, the gains are substantial and correlated with the number of skipped layers, with a $4\times$ speedup on SmolVLA and up to $3.6\times$ on $\pi_{0.5}$.
Notably, as demonstrated by the negligible variation in $\Delta$SR, the addition of the KV cache synthesis mechanism does not affect the success rate.
Furthermore, we observe two different trends in the compute savings.
For instance, on SmolVLA the prefix is a small part of the
total compute ($86.3$ of $559$ GFLOPs), so even a $4\times$ prefix reduction is worth $11.6\%$ at the end.
On the other hand, the prefix dominates on $\pi_{0.5}$ ($3939$ of $5213$ GFLOPs) and the same mechanism removes up to
$54.6\%$ of the total compute. The synthesis is therefore the component that makes the $V$ axis a compute lever on large backbones.

\section{CONCLUSIONS}
We presented a training framework for pretrained flow-matching VLAs that allows us to study the depth (i.e., the number of layers) of the VLM backbone $V$, the depth of the action expert $A$, and the number of denoising steps $D$ as three jointly configurable compute axes.
Exit Transformers distilled from the frozen policy make intermediate depths valid exits in the original action space, and a KV cache synthesis mechanism fills the prefix cache of the skipped backbone layers, allowing the action expert to exit deeper than the backbone.
Across two policies (i.e., SmolVLA and $\pi_{0.5}$) and two benchmarks (i.e., LIBERO and Meta-World), the three axes proved complementary: $V$ reduces FLOPs, $A$ affects latency, and $D$ improves both, with their joint configuration dominating the full policy and the single axis on the Pareto frontier.
Our task analysis further revealed that no fixed compute budget is optimal across tasks and that each task is sensitive to different compute axis in $(V, A, D)$, with intermediate distilled representations potentially outperforming the full model.

Nevertheless, some limitations remain. For instance, early exits are attached at specific depths, leaving open the question of what the optimal set of depths is for each policy.

Finally, as showcased by our task analysis, VLAs should learn to flexibly allocate compute budget to improve efficiency and performance. Future work should investigate adaptive compute selection strategies, able to tailor compute allocation to each timestep of an episode or across task types, with the potential to further improve the Pareto frontier.

\bibliographystyle{IEEEtran}
\bibliography{bibliography}

@IEEEtranBSTCTL{BSTcontrol,
    CTLuse_forced_etal  = "yes",
    CTLmax_names_forced_etal = "5",
    CTLnames_show_etal = "5"
}

@InProceedings{zitkovich2023rt,
  title = 	 {RT-2: Vision-Language-Action Models Transfer Web Knowledge to Robotic Control},
  author =       {Zitkovich, Brianna and Yu, Tianhe and Xu, Sichun and Xu, Peng and Xiao, Ted and Xia, Fei and Wu, Jialin and Wohlhart, Paul and Welker, Stefan and Wahid, Ayzaan and Vuong, Quan and Vanhoucke, Vincent and Tran, Huong and Soricut, Radu and Singh, Anikait and Singh, Jaspiar and Sermanet, Pierre and Sanketi, Pannag R. and Salazar, Grecia and Ryoo, Michael S. and Reymann, Krista and Rao, Kanishka and Pertsch, Karl and Mordatch, Igor and Michalewski, Henryk and Lu, Yao and Levine, Sergey and Lee, Lisa and Lee, Tsang-Wei Edward and Leal, Isabel and Kuang, Yuheng and Kalashnikov, Dmitry and Julian, Ryan and Joshi, Nikhil J. and Irpan, Alex and Ichter, Brian and Hsu, Jasmine and Herzog, Alexander and Hausman, Karol and Gopalakrishnan, Keerthana and Fu, Chuyuan and Florence, Pete and Finn, Chelsea and Dubey, Kumar Avinava and Driess, Danny and Ding, Tianli and Choromanski, Krzysztof Marcin and Chen, Xi and Chebotar, Yevgen and Carbajal, Justice and Brown, Noah and Brohan, Anthony and Arenas, Montserrat Gonzalez and Han, Kehang},
  booktitle = 	 {Proceedings of The 7th Conference on Robot Learning},
  pages = 	 {2165--2183},
  year = 	 {2023},
  editor = 	 {Tan, Jie and Toussaint, Marc and Darvish, Kourosh},
  volume = 	 {229},
  series = 	 {Proceedings of Machine Learning Research},
  month = 	 {06--09 Nov},
  publisher =    {PMLR}
}

@article{kawaharazuka2025vla-survey,
  title={Vision-language-action models for robotics: A review towards real-world applications},
  author={Kawaharazuka, Kento and Oh, Jihoon and Yamada, Jun and Posner, Ingmar and Zhu, Yuke},
  journal={IEEE Access},
  year={2025},
  publisher={IEEE}
}

@article{doi:10.1177/02783649261468360,
author = {Xiangtong Yao and Hongkuan Zhou and Oier Mees and Yuan Meng and Ted Xiao and Yonatan Bisk and Jean Oh and Edward Johns and Mohit Shridhar and Dhruv Shah and Jesse Thomason and Kai Huang and Joyce Chai and Zhenshan Bing and Alois Knoll},
title ={Bridging language and action: A survey of language-conditioned robot manipulation},
journal = {The International Journal of Robotics Research},
year = {2026},
doi = {10.1177/02783649261468360},
}

@InProceedings{kim2024openvla,
  title = 	 {OpenVLA: An Open-Source Vision-Language-Action Model},
  author =       {Kim, Moo Jin and Pertsch, Karl and Karamcheti, Siddharth and Xiao, Ted and Balakrishna, Ashwin and Nair, Suraj and Rafailov, Rafael and Foster, Ethan P and Sanketi, Pannag R and Vuong, Quan and Kollar, Thomas and Burchfiel, Benjamin and Tedrake, Russ and Sadigh, Dorsa and Levine, Sergey and Liang, Percy and Finn, Chelsea},
  booktitle = 	 {Proceedings of The 8th Conference on Robot Learning},
  pages = 	 {2679--2713},
  year = 	 {2025},
  editor = 	 {Agrawal, Pulkit and Kroemer, Oliver and Burgard, Wolfram},
  volume = 	 {270},
  series = 	 {Proceedings of Machine Learning Research},
  month = 	 {06--09 Nov},
  publisher =    {PMLR}
}

@inproceedings{black2024pi0,
  title = {$\pi_0$: A Vision-Language-Action Flow Model for General Robot Control},
  author = {Kevin Black and Noah Brown and Danny Driess and Adnan Esmail and Michael Robert Equi and Chelsea Finn and Niccolo Fusai and Lachy Groom and Karol Hausman and Brian Ichter and Szymon Jakubczak and Tim Jones and Liyiming Ke and Sergey Levine and Adrian Li-Bell and Mohith Mothukuri and Suraj Nair and Karl Pertsch and Lucy Xiaoyang Shi and Laura Smith and James Tanner and Quan Vuong and Anna Walling and Haohuan Wang and Ury Zhilinsky},
  booktitle = {RSS 2025},
  year = {2025}
}

@article{shukor2025smolvla,
  title={Smolvla: A vision-language-action model for affordable and efficient robotics},
  author={Shukor, Mustafa and Aubakirova, Dana and Capuano, Francesco and Kooijmans, Pepijn and Palma, Steven and Zouitine, Adil and others},
  journal={arXiv preprint arXiv:2506.01844},
  year={2025}
}

@InProceedings{intelligence2025pi05,
  title = 	 {$\pi_{0.5}$: a Vision-Language-Action Model with Open-World Generalization},
  author =       {Black, Kevin and Brown, Noah and Darpinian, James and Dhabalia, Karan and Driess, Danny and Esmail, Adnan and Equi, Michael Robert and Finn, Chelsea and Fusai, Niccolo and Galliker, Manuel Y. and Ghosh, Dibya and Groom, Lachy and Hausman, Karol and ichter, brian and Jakubczak, Szymon and Jones, Tim and Ke, Liyiming and LeBlanc, Devin and Levine, Sergey and Li-Bell, Adrian and Mothukuri, Mohith and Nair, Suraj and Pertsch, Karl and Ren, Allen Z. and Shi, Lucy Xiaoyang and Smith, Laura and Springenberg, Jost Tobias and Stachowicz, Kyle and Tanner, James and Vuong, Quan and Walke, Homer and Walling, Anna and Wang, Haohuan and Yu, Lili and Zhilinsky, Ury},
  booktitle = 	 {Proceedings of The 9th Conference on Robot Learning},
  pages = 	 {17--40},
  year = 	 {2025},
  editor = 	 {Lim, Joseph and Song, Shuran and Park, Hae-Won},
  volume = 	 {305},
  series = 	 {Proceedings of Machine Learning Research},
  month = 	 {27--30 Sep},
  publisher =    {PMLR}
}

@article{wen2025tinyvla,
  title={TinyVLA: toward fast, data-efficient vision-language-action models for robotic manipulation},
  author={Wen, Junjie and Zhu, Yichen and Li, Jinming and Zhu, Minjie and Tang, Zhibin and Wu, Kun and Xu, Zhiyuan and Liu, Ning and Cheng, Ran and Shen, Chaomin and others},
  journal={IEEE Robotics and Automation Letters},
  volume={10},
  number={4},
  pages={3988--3995},
  year={2025},
  publisher={IEEE}
}

@inproceedings{laskaridis2021adaptive,
  title={Adaptive inference through early-exit networks: Design, challenges and directions},
  author={Laskaridis, Stefanos and Kouris, Alexandros and Lane, Nicholas D},
  booktitle={Proceedings of the 5th International Workshop on Embedded and Mobile Deep Learning},
  pages={1--6},
  year={2021}
}

@inproceedings{teerapittayanon2016branchynet,
  title={Branchynet: Fast inference via early exiting from deep neural networks},
  author={Teerapittayanon, Surat and McDanel, Bradley and Kung, Hsiang-Tsung},
  booktitle={2016 23rd international conference on pattern recognition (ICPR)},
  pages={2464--2469},
  year={2016},
  organization={IEEE}
}

@inproceedings{xin2020deebert,
  title={DeeBERT: Dynamic early exiting for accelerating BERT inference},
  author={Xin, Ji and Tang, Raphael and Lee, Jaejun and Yu, Yaoliang and Lin, Jimmy},
  booktitle={Proceedings of the 58th annual meeting of the association for computational linguistics},
  pages={2246--2251},
  year={2020}
}

@article{yue2024deer,
  title={Deer-vla: Dynamic inference of multimodal large language models for efficient robot execution},
  author={Yue, Yang and Wang, Yulin and Kang, Bingyi and Han, Yizeng and Wang, Shenzhi and Song, Shiji and Feng, Jiashi and Huang, Gao},
  journal={Advances in Neural Information Processing Systems},
  volume={37},
  pages={56619--56643},
  year={2024}
}

@article{xu2026vla,
  title={Vla-cache: Efficient vision-language-action manipulation via adaptive token caching},
  author={Xu, Siyu and Wang, Yunke and Xia, Chenghao and Zhu, Dihao and Huang, Tao and Xu, Chang},
  journal={Advances in Neural Information Processing Systems},
  volume={38},
  pages={164448--164473},
  year={2026}
}

@article{yang2026efficientvla,
  title={Efficientvla: Training-free acceleration and compression for vision-language-action models},
  author={Yang, Yantai and Wang, Yuhao and Wen, Zichen and Zhongwei, Luo and Zou, Chang and Zhang, Zhipeng and Wen, Chuan and Zhang, Linfeng},
  journal={Advances in Neural Information Processing Systems},
  volume={38},
  pages={40891--40914},
  year={2026}
}

@article{song2025ceed,
  title={Ceed-vla: Consistency vision-language-action model with early-exit decoding},
  author={Song, Wenxuan and Chen, Jiayi and Ding, Pengxiang and Huang, Yuxin and Zhao, Han and Wang, Donglin and Li, Haoang},
  journal={arXiv preprint arXiv:2506.13725},
  year={2025}
}

@inproceedings{zhang2026mole,
  title={Mole-vla: Dynamic layer-skipping vision language action model via mixture-of-layers for efficient robot manipulation},
  author={Zhang, Rongyu and Dong, Menghang and Zhang, Yuan and Heng, Liang and Chi, Xiaowei and Dai, Gaole and Du, Li and Wang, Dan and Du, Yuan and Zhang, Shanghang},
  booktitle={Proceedings of the AAAI Conference on Artificial Intelligence},
  volume={40},
  number={22},
  pages={18764--18772},
  year={2026}
}

@article{luan2026snapflow,
  title={Snapflow: One-step action generation for flow-matching vlas via progressive self-distillation},
  author={Luan, Wuyang and Li, Junhui and Zhao, Weiguang and Zhang, Wenjian and Wu, Tieru and Ma, Rui},
  journal={arXiv preprint arXiv:2604.05656},
  year={2026}
}

@article{yu2026ac,
  title={AC\^{} 2-VLA: Action-Context-Aware Adaptive Computation in Vision-Language-Action Models for Efficient Robotic Manipulation},
  author={Yu, Wenda and Wang, Tianshi and Li, Fengling and Li, Jingjing and Zhu, Lei},
  journal={arXiv preprint arXiv:2601.19634},
  year={2026}
}

@article{zhang2026a1,
  title={A1: A fully transparent open-source, adaptive and efficient truncated vision-language-action model},
  author={Zhang, Kaidong and Zhang, Jian and Xu, Rongtao and Sun, Yu and Xue, Shuoshuo and Wen, Youpeng and Guo, Xiaoyu and Guo, Minghao and Liufu, Weijia and Zihou, Liu and others},
  journal={arXiv preprint arXiv:2604.05672},
  year={2026}
}

@inproceedings{lipman2023flow,
title={Flow Matching for Generative Modeling},
author={Yaron Lipman and Ricky T. Q. Chen and Heli Ben-Hamu and Maximilian Nickel and Matthew Le},
booktitle={The Eleventh International Conference on Learning Representations },
year={2023},
url={https://openreview.net/forum?id=PqvMRDCJT9t}
}

@inproceedings{elbayad2020depth,
title={Depth-Adaptive Transformer},
author={Maha Elbayad and Jiatao Gu and Edouard Grave and Michael Auli},
booktitle={International Conference on Learning Representations},
year={2020},
url={https://openreview.net/forum?id=SJg7KhVKPH}
}

@article{schuster2022confident,
  title={Confident adaptive language modeling},
  author={Schuster, Tal and Fisch, Adam and Gupta, Jai and Dehghani, Mostafa and Bahri, Dara and Tran, Vinh and Tay, Yi and Metzler, Donald},
  journal={Advances in Neural Information Processing Systems},
  volume={35},
  pages={17456--17472},
  year={2022}
}

@inproceedings{elhoushi2024layerskip,
  title={Layerskip: Enabling early exit inference and self-speculative decoding},
  author={Elhoushi, Mostafa and Shrivastava, Akshat and Liskovich, Diana and Hosmer, Basil and Wasti, Bram and Lai, Liangzhen and Mahmoud, Anas and Acun, Bilge and Agarwal, Saurabh and Roman, Ahmed and others},
  booktitle={Proceedings of the 62nd Annual Meeting of the Association for Computational Linguistics (Volume 1: Long Papers)},
  pages={12622--12642},
  year={2024}
}

@inproceedings{bajpai2025free,
  title={FREE: fast and robust vision language models with early exits},
  author={Bajpai, Divya Jyoti and Hanawal, Manjesh Kumar},
  booktitle={Findings of the Association for Computational Linguistics: ACL 2025},
  pages={23599--23615},
  year={2025}
}

@inproceedings{cadene2026lerobot,
  title={Lerobot: An open-source library for end-to-end robot learning},
  author={Cadene, Remi and Alibert, Simon and Capuano, Francesco and Aractingi, Michel and Zouitine, Adil and Kooijmans, Pepijn and Choghari, Jade and Russi, Martino and Pascal, Caroline and Palma, Steven and others},
  booktitle={International Conference on Learning Representations},
  volume={2026},
  pages={122398--122417},
  year={2026}
}

@article{liu2023libero,
  title={Libero: Benchmarking knowledge transfer for lifelong robot learning},
  author={Liu, Bo and Zhu, Yifeng and Gao, Chongkai and Feng, Yihao and Liu, Qiang and Zhu, Yuke and Stone, Peter},
  journal={Advances in Neural Information Processing Systems},
  volume={36},
  pages={44776--44791},
  year={2023}
}

@inproceedings{yu2020meta,
  title={Meta-world: A benchmark and evaluation for multi-task and meta reinforcement learning},
  author={Yu, Tianhe and Quillen, Deirdre and He, Zhanpeng and Julian, Ryan and Hausman, Karol and Finn, Chelsea and Levine, Sergey},
  booktitle={Conference on robot learning},
  pages={1094--1100},
  year={2020},
  organization={PMLR}
}

\end{document}